# Simultaneous Policy Learning and Latent State Inference for Imitating Driver Behavior

Jeremy Morton[1] and Mykel J. Kochenderfer[1]

*Abstract*— Human driving depends on latent states, such as aggression and intent, that cannot be directly observed. In this work, we propose a method for learning driver models that can account for unobserved states. When trained on a synthetic dataset, our model is able to learn encodings for vehicle trajectories that distinguish between four distinct classes of driver behavior. Such encodings are learned without any knowledge of the number of driver classes or any objective that directly requires the model to learn encodings for each class. We show that latent state driving policies outperform baseline methods at replicating driver behavior. Furthermore, we demonstrate that the actions chosen by our policy are heavily influenced by its assigned latent state.

## I. INTRODUCTION

Due to their low cost, evaluation time, and risk compared to real-world driving tests, simulations will play a key role in the advancement and acceptance of future automated driving systems. For such simulations to provide accurate estimates of system performance, they will require representative models for human driver behavior. While much of the early work in human driver modeling has focused on constructing rule-based controllers for approximating human driver behavior [1]–[3], data-driven approaches are now gaining traction due to increased data availability and greater model flexibility.

A common data-driven approach is behavioral cloning. This approach involves training models to maximize the likelihood assigned to actions taken by human drivers in a training dataset [4]–[7]. Behavioral cloning models can provide accurate action predictions over short time horizons, but ultimately their performance tends to suffer from cascading errors, where small errors in action predictions lead vehicles into collisions or other scenarios often not covered by training data [8].

Recent data-driven approaches have focused on applying techniques from the sequential decision making field to the task of imitating human drivers [9], [10]. Such reward-based learning techniques can propagate blame for undesirable outcomes back to all preceding actions, and hence avoid the pitfalls associated with the short-sighted action selection of behavioral cloning approaches. In particular, recent work has focused on learning neural network policies through generative adversarial imitation learning (GAIL) [11], and demonstrated the ability to learn data-driven human driver models capable of generating long-term trajectories that are largely unplagued by cascading errors [12].

[1]Department of Aeronautics and Astronautics at Stanford University, Stanford, CA 94305, USA {jmorton2, mykel}@stanford.edu

Beyond exhibiting stable behavior, human driver models should aim to generate actions that are representative of the intricacies and variety of behaviors that can be found in real-world human driving. Intuitively, unobserved (latent) states such as driver style, driver intention, or maneuver classes have a large influence over the actions that drivers select, and thus driver models should be capable of accounting for such states. Furthermore, latent state inference can prove valuable for tasks aside from human driver modeling. For example, knowledge of latent driver styles in surrounding vehicles can improve safety in planning for freeway driving [13]. Prior work in latent state inference in driving contexts has employed a broad range of techniques such as fuzzy logic, hidden Markov models, clustering, and deep learning [14]–[17]. However, these approaches require assumptions about the number or meaning of latent states that may not prove to be valid on naturalistic driving datasets.

Building upon recent advances in variational inference and deep learning, we propose a method that learns an encoder for inferring vehicle latent states while simultaneously learning a policy that accounts for such latent states in selecting driver actions. By using a real-valued latent state representation, our approach makes no assumptions about the meaning or number of latent states. Trained on a synthetic dataset, our proposed method is shown to produce encoders that distinguish between four distinct driver classes, along with policies that modify their action outputs depending on their assigned latent state, thereby allowing them to generate trajectories that better capture the variety of driving behaviors exhibited within the training data.

## II. PROBLEM FORMULATION

Variational inference is a machine learning technique for approximating unknown and often intractable probability densities [18]. In this work, we apply variational inference to the task of learning driver policies, where we assume that there is a latent state that influences the actions that drivers take. We assume that drivers select actions based on both their current roadway state $s$ and the unobserved latent state $z$. While latent states may change over long time periods, in this work we assume $z$ is constant throughout the trajectory of interest. A dynamic Bayesian network (DBN) illustrating the structure for this problem is shown in Fig. 1.

In learning a driving policy, the standard behavioral cloning objective is to maximize $\frac{1}{T} \sum_{t=1}^{T} \log p(a_t \mid s_t)$, the probability a model assigns to actions taken by experts throughout a trajectory. Maximizing this objective directly overlooks the effect that latent states may have on driver

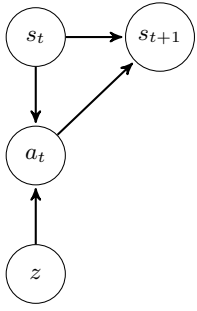

Fig. 1: Dynamic Bayesian network governing time evolution of a vehicle's roadway state. It is assumed drivers select actions based on their current state $s_t$ and some unobserved state $z$.

actions. Considering a single component of the behavioral cloning objective, we note that

$$\log p(a_t \mid s_t) = \log \int p(a_t, z \mid s_t) dz \quad (1)$$

$$= \log \int p(a_t \mid z, s_t) p(z \mid s_t) dz. \quad (2)$$

We assume that $z$ is constant in time, so $p(z \mid s_t) = p(z)$, our prior over $z$. Suppose we have a sequence of states $\boldsymbol{s}$ and actions $\boldsymbol{a}$ and want to infer the value of $z$. The true posterior distribution $p(z \mid \boldsymbol{a}, \boldsymbol{s})$ is unknown and is generally intractable. Hence, we introduce $q(z \mid \boldsymbol{a}, \boldsymbol{s})$, an approximation to the true posterior $p(z \mid \boldsymbol{a}, \boldsymbol{s})$. Incorporating this distribution into Eq. (2), we obtain

$$\log p(a_t \mid s_t) = \log \int p(a_t \mid z, s_t) p(z) \frac{q(z \mid \boldsymbol{a}, \boldsymbol{s})}{q(z \mid \boldsymbol{a}, \boldsymbol{s})} dz \quad (3)$$

$$= \log \mathbb{E}_{z \sim q} \left[ \frac{p(a_t \mid z, s_t) p(z)}{q(z \mid \boldsymbol{a}, \boldsymbol{s})} \right]. \quad (4)$$

Now invoking Jensen's inequality, we find that

$$\log p(a_t \mid s_t) \geq \mathbb{E}_{z \sim q} \left[ \log p(a_t \mid z, s_t) \right] -$$
$$\mathcal{D}_{\text{KL}} \left[ q(z \mid \boldsymbol{a}, \boldsymbol{s}) \,\|\, p(z) \right]. \quad (5)$$

Applying this result to state-action pairs for an entire trajectory, we find a lower bound for the behavioral cloning objective:

$$\frac{1}{T} \sum_{t=1}^{T} \log p(a_t \mid s_t) \geq \mathbb{E}_{z \sim q} \left[ \frac{1}{T} \sum_{t=1}^{T} \log p(a_t \mid z, s_t) \right] -$$
$$\mathcal{D}_{\text{KL}} \left[ q(z \mid \boldsymbol{a}, \boldsymbol{s}) \,\|\, p(z) \right]. \quad (6)$$

Thus, if we we can evaluate and improve the expression on the right-hand side, then we can raise the lower bound on our objective.

## III. Training

We now discuss how to train driver policies that select actions conditioned on the current roadway state *and* a latent state that is constant throughout time. First, we describe how we represent our policies, and then we derive and explain the objective function used to train these policies.

### A. Policy Representation

We represent all policies using neural networks due to their ability to serve as universal function approximators that automatically extract relevant features from complex, high-dimensional inputs. The networks in this work take on two forms: (1) multilayer perceptrons (MLPs), which are feedforward networks that manipulate inputs through a series of matrix multiplications and element-wise activations and (2) recurrent neural networks, which maintain an internal state that allows them to generate outputs conditioned on entire sequences of inputs.

We use rectified linear unit (ReLU) activations [19] in our multilayer perceptrons and long short-term memory (LSTM) cells [20] in our recurrent networks. All neural networks consist of two hidden layers with 128 units in each layer. Our policies are stochastic, which means that each policy must map its input to a distribution over acceleration and turn-rate values. We represent this distribution as a multivariate Gaussian, where the network outputs the mean and variance values for each action, and a diagonal covariance matrix is assumed, as is common in deep learning [21].

### B. Optimization Procedure

We now derive a practical training procedure. Our goal is to train a policy through a process that raises the lower bound on the right-hand side of Eq. (6). Note that if we represent our policy $\pi$ by a multilayer perceptron with parameters $\boldsymbol{\theta}$, then we have $p(a_t \mid s_t, z) = \pi_{\boldsymbol{\theta}}(a_t \mid s_t, z)$. We can set our prior over $z$ to be $p(z) = \mathcal{N}(0, I)$. Finally, the question remains as to how to compute the approximate posterior over $z$ based on the entire sequence of state and action values. If we represent $q$ using a recurrent neural network with parameters $\boldsymbol{\phi}$, then the hidden state at time $T$ is given by $h_T = f_{\boldsymbol{\phi}}(s_T, a_T, h_{T-1})$, encoding information about the states and actions at all time steps up to time $T$. We therefore replace $q(z \mid \boldsymbol{a}, \boldsymbol{s})$ with $q_{\boldsymbol{\phi}}(z \mid h_T)$. By forming a Monte Carlo estimate of the expectation in Eq. (6), we arrive at the following loss function:

$$\mathcal{L} = -\frac{1}{LT} \sum_{\ell=1}^{L} \sum_{t=1}^{T} \log \pi_{\boldsymbol{\theta}}(a_t \mid s_t, z^{(\ell)})$$
$$+ \lambda \mathcal{D}_{\text{KL}} \left[ q_{\boldsymbol{\phi}}(z \mid h_T) \,\|\, p(z) \right], \quad (7)$$

where the $z^{(\ell)}$'s are samples drawn from the approximation to the posterior and $\lambda$ is a hyperparameter that balances the two components of the loss function.

An illustration of the encoder, policy, and optimization procedure can be found in Fig. 2. For training, we let $T = 50$ and $L = 10$. In experiments, the dimensionality of the $z$-distribution did not have a strong influence on policy performance, with a one-, two-, and three-dimensional Gaussians achieving comparable results. For ease of visualization, the distribution over $z$ is chosen to be two-dimensional with diagonal covariance. The parameters to this approximate posterior distribution are output by the encoder network, much like the action distribution output by the policy. Finally, we require that $\lambda \ll 1$ or else the gains made by driving the approximate posterior closer to the prior far outweigh the gains that can be made by improving the policy. In our experiments, we gradually increase the value of $\lambda$ throughout training until it reaches a value of $\lambda = 0.05$.

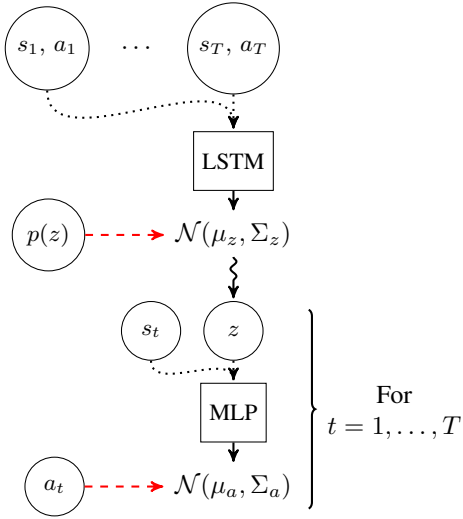

Fig. 2: Visualization of networks and training process. Red arrows represent relationships relevant to the optimization objective.

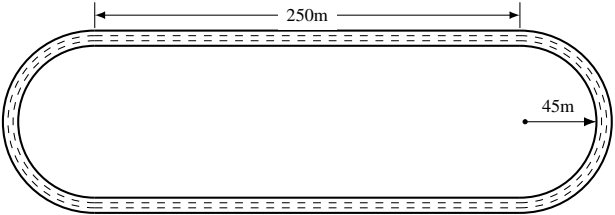

Fig. 3: Roadway used in simulation environment.

## IV. DRIVING ENVIRONMENT

Our policy is trained on a synthetic dataset and then evaluated in the same simulation environment used to generate the training data. The following sections describe this simulation environment.

### A. Roadway

All driving simulations were run on an oval racetrack, as illustrated in Fig. 3. The roadway consists of two 250 m straightaways and two 45 m radius turns. In generating the training dataset, cars are initialized three-across at 11 evenly spaced locations with a separation of 75 m, for a total of 33 vehicles. Simulations are carried out for 30 s.

### B. Driver Classes

All vehicles within the environment use a rule-based controller to select their actions. Longitudinal acceleration and braking actions are governed by the intelligent driver model (IDM) [2], lane change decisions are governed by MOBIL [3], and a proportional controller is used for lane tracking. To arrive at four distinct classes of driver behavior, each class is given a unique IDM parameterization. The IDM parameters that are varied are: desired speed ($v_{\text{des}}$), minimum headway distance ($d_{\min}$), desired time headway ($\tau$), maximum acceleration ($a_{\max}$), and comfortable deceleration ($b_{\text{comf}}$). The parameter settings for each class can be found in Table I.

TABLE I: IDM Parameter Settings

| | $v_{\text{des}}$ (m/s) | $d_{\min}$ (m) | $\tau$ (s) | $a_{\max}$ (m/s$^2$) | $b_{\text{comf}}$(m/s$^2$) |
|---|---|---|---|---|---|
| Passive | 10 | 5 | 1.75 | 1 | 1 |
| Aggressive | 30 | 1 | 0.25 | 5 | 5 |
| Tailgater | 15 | 1 | 0.25 | 1 | 1 |
| Speeder | 30 | 5 | 1.75 | 5 | 5 |

*Passive* drivers have a low desired speed and prefer large headway distances and small accelerations. In contrast, *aggressive* drivers have a preference for faster speeds, and are comfortable with smaller headway distances and larger accelerations. The two additional driver classes share characteristics with both the *passive* and *aggressive* drivers. *Tailgaters* share the preference of *passive* drivers for low speeds and accelerations, but are willing to drive closer to the vehicle in front of them, and they are given a desired speed slightly higher than that of *passive* drivers to allow them to exhibit these tailgating characteristics. *Speeders* share a tolerance for high speeds and accelerations with *aggressive* drivers, but are hesitant to travel too closely to the vehicle in front of them. To populate the simulation environment, the class of each vehicle is sampled at random. All IDM parameters are set according to the values in Table I with the exception of desired speed, which is subject to zero-mean Gaussian noise of unit variance.

### C. Features

We extract a set of features to summarize information about the local roadway scene surrounding a given vehicle. For each car, we can extract a set of 40 LIDAR-like features, which consist of equally spaced beams spanning the full 360 degrees around the vehicle. Twenty of these beams capture range information about the vehicles that they intersect, while the remaining 20 beams measure range rate. As in previous work, these LIDAR-like features are augmented with an additional set of features that capture information about the ego vehicle's shape, odometry, and lane-relative position [12]. In total, 48 features are used to represent the vehicle state, and these extracted features are fed into the neural policies during training and rollouts.

## V. EVALUATION

Learned models were evaluated in three stages. First, we examine the latent encodings that are learned for trajectories corresponding to each driver style. Next, we study how accurate our policy is in replicating the driver behavior in our synthetic dataset. Finally, we explore how variations in the latent variable values fed into the policy influence the actions that it outputs.

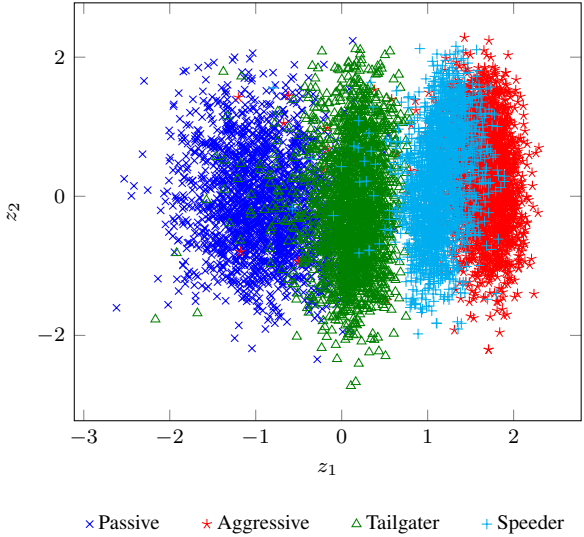

Fig. 4: Visualization of driver style encodings in a two-dimensional latent space.

## A. Latent Space Encodings

Once we have a trained model, we can run trajectories through the encoder to observe how it represents the trajectories within the space of latent encodings. To perform this analysis, we extract a dataset for each driver class from simulations where all vehicles on the roadway are assigned the same class. Due to random variations in desired speed, these datasets will still exhibit the tailgating tendencies for each class but will avoid situations where, for example, an aggressive driver is trapped behind a passive driver, and hence drives much slower than its desired speed. We then feed trajectories from these datasets through the encoder and sample $z$-values from the output distribution.

The result of following this procedure for each dataset and plotting the latent state samples can be found in Fig. 4. We can see that the encodings corresponding to each class are largely distinct. In particular, there is a clear separation between the *aggressive/speeder* classes and the *passive/tailgater* classes, illustrating that the encoder learns it is crucial that $z$-values convey information about the magnitude of speed and acceleration values. However, the fact that there is further separation between the classes shows that the encoder also learns to capture information about the tailgating behavior of each vehicle. It is worth stressing that the encoder was trained on an unlabeled dataset and is never told how many driver classes there are or explicitly instructed to find distinct encodings for each class. Rather, it learns distinct encodings for each driver class because doing so is *useful* to the policy, as discussed in the next section.

## B. Policy Performance

*1) Experiments:* To understand the quality of our learned policy, we conduct rollouts with the policy and extract metrics to quantify its performance relative to three baseline models. The first baseline model is an MLP policy that receives only the current state as input and is trained to maximize the standard behavioral cloning objective $p(a_t \mid s_t)$ without considering any possible latent states. A second baseline model attempts to maximize the same objective as the MLP policy, but is represented by an LSTM network. It receives 5 s of vehicle state values to initialize its internal state prior to generating action predictions, and hence may be able to make inferences about driver style even though it does not explicitly attempt to generate latent state encodings. The final baseline model, deemed the Oracle, is designed to provide an upper bound on performance for policies with knowledge of latent states. The Oracle model is an MLP that receives the vehicle state concatenated with a one-hot vector that communicates the true driver style of the vehicle that generated each training example.

All models we consider here are behavioral cloning policies, and hence suffer from cascading errors. In particular, if these policies encounter states during rollouts that are not present in the training data, such as off-road driving, they may take extreme actions that make it difficult to accurately evaluate their performance. Thus, we only perform rollouts for vehicles driving on the straight sections of the roadway, and we set the turn-rate to zero throughout the entire simulation. Since the different driver classes only differ in their longitudinal driving behavior, this evaluation procedure still allows us to quantify the effectiveness of each model at producing accurate action predictions for each class, while largely mitigating the problems associated with cascading errors.

Our evaluation is performed on a test set of 1000 five-second trajectories running at 10 Hz. For each trajectory, we perform 20 rollouts with each model and quantify how closely these rollouts recreate the behavior inherent in the trajectory. In each rollout with our latent-aware policy, all vehicle states and actions from the trajectory are first run through the encoder to generate latent state samples that can be provided to the policy. In each rollout with the LSTM policy, the previous 5 s worth of vehicle state values are first fed into the network to initialize its internal state. The MLP and Oracle policies are both reactive, and hence do not require that any special procedures occur prior to performing rollouts.

To quantify model performance, we first extract the root-weighted square error (RWSE) for displacement and speed over various prediction horizons. The root-weighted square error finds the square error between a predicted and true value, and weights the error according to the probability mass associated with the true value [22]. A Monte Carlo estimate for the RWSE in variable $v$ at time horizon $H$ is given by

$$\text{RWSE}_H = \sqrt{\frac{1}{mn} \sum_{i=1}^{m} \sum_{j=1}^{n} \left( v_H^{(i)} - \hat{v}_H^{(i,j)} \right)^2}, \quad (8)$$

where $m$ is the number of trajectories, $n$ is the number of simulated traces per trajectory, $v_H^{(i)}$ is the true value of the variable in trajectory $i$, and $\hat{v}_H^{(i,j)}$ is the value of the variable in rollout $j$ associated with trajectory $i$. Since our vehicles are constrained to remain in their lane, displacement is measured longitudinally in the Frenet frame.

TABLE II: KL-Divergence Values

|            | iTTC                  | Speed                 | Acceleration |
|------------|-----------------------|-----------------------|--------------|
| Latent-MLP | $6.42\times10^{-2}$   | $2.61\times10^{-2}$   | 0.235        |
| MLP        | $3.90\times10^{-2}$   | $2.12\times10^{-2}$   | 0.408        |
| LSTM       | $2.73\times10^{-2}$   | $3.06\times10^{-2}$   | **0.201**    |
| Oracle     | $\mathbf{1.25 \times 10^{-2}}$ | $\mathbf{1.73 \times 10^{-2}}$ | 0.227        |

In addition to RWSE, we consider how closely the empirical distribution over emergent quantities in policy rollouts matches the distributions over the same quantities in test-set trajectories. We use the KL-divergence to quantify the similarity between distributions over speed, acceleration, and inverse time-to-collision (iTTC). The empirical distributions are modeled as piecewise uniform with 100 bins of equal width.

*2) Results:* The KL-divergence values can be found in Table II and the RWSE values can be found in Fig. 5, with results for our model labeled Latent-MLP. All models achieve low KL-divergence in iTTC and speed, with the Oracle slightly outperforming the other policies. Furthermore, we can see that the distribution over acceleration values in MLP trajectories does not match the true distribution as well as the acceleration distributions in trajectories generated by the other models. These other models either have access to a series of previous vehicle states or a latent state, which provides them with indirect information about vehicle acceleration preferences that the MLP lacks, and may explain its relatively poor performance.

We can see the clear benefit of policy awareness about latent states in the RWSE results. Both the Oracle and our Latent-MLP model achieve lower error than the other models in speed prediction, especially over longer prediction horizons. Given that the latent states in essence provide policies with *a priori* information about the desired speed for each vehicle, it makes sense that policies with knowledge of driver style would make more accurate speed predictions, which in turn lead to more accurate position predictions.

*C. Latent Variable Settings*

One appealing outcome of training a policy with this approach is that, if the set of learned encodings approximately matches the prior distribution, then we can populate a roadway scene with vehicles driven by our policy and sample a latent state from the prior $z \sim p(z)$ for each vehicle. If the policy has learned to act differently depending on the latent variable it receives as an input, then such a procedure will effectively assign a different behavior to each vehicle on the roadway. We now study the effect that different latent variable settings have on the actions selected by our policy. Because Fig. 4 provides us with interpretable latent encodings for each of our driver classes, for these experiments we let $z_1 \in \{-1, 0, 1.5, 2\}$ and set $z_2 = 0$.

In a first set of experiments, we perform 100 five-second rollouts with vehicles assigned each of the latent variable settings. As in the rollouts performed to measure policy performance, rollouts are performed on straightaways and all turn-rate values are set to zero. Initial vehicle speeds are drawn from a normal distribution centered at 20 m/s, and no other vehicles are initialized to be in front of the ego vehicle. We then track how the distribution over vehicle speeds varies as time progresses, the results of which can be seen in the top row of Fig. 6 for horizons of 0, 1, and 5 seconds. From the plots, we can see that the latent variable setting has a significant effect on the resulting speed of the vehicle. In fact, the resulting speeds for each latent state value are remarkably close to the desired speed for the corresponding driver class: *passive* drivers end up traveling at about 10 m/s, *tailgaters* end up traveling at about 15 m/s, and *speeders* and *aggressive* drivers both travel at close to 30 m/s, with *aggressive* drivers moving slightly faster.

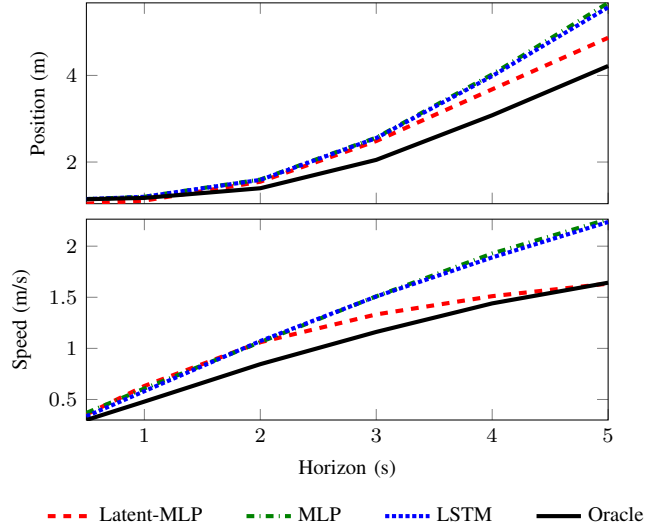

Fig. 5: Root weighted square error in position and velocity values over various prediction horizons.

In a second set of experiments, we study the tailgating behavior of vehicles with different latent state settings. As before, we perform 100 rollouts and initialize vehicles with speeds close to 20 m/s. However, in this case we also initialize a vehicle with a desired speed of 20 m/s on the roadway 8 m in front of the ego vehicle. Then, in the resulting rollouts, we can observe the distribution over headway distances with each latent state setting, as shown in the lower three plots of Fig. 6. We observe that vehicles assigned an *aggressive* latent state end up driving much closer to the vehicle in front of them than vehicles assigned a latent state corresponding to the *speeder* driving class. Additionally, in all cases the vehicles do not accelerate all the way to their desired speed, which would cause them to collide with the vehicle in front of them.

VI. CONCLUSIONS

We introduced a method for jointly learning an encoder capable of inferring latent states and a policy that selects actions with awareness of such unobserved states. With an unlabeled dataset, our method is able to learn distinct encodings for four different driver classes. Furthermore, our policy learns to select different actions depending on the setting of its latent state, which allows it to perform rollouts that are more accurate and representative of different driver behaviors.

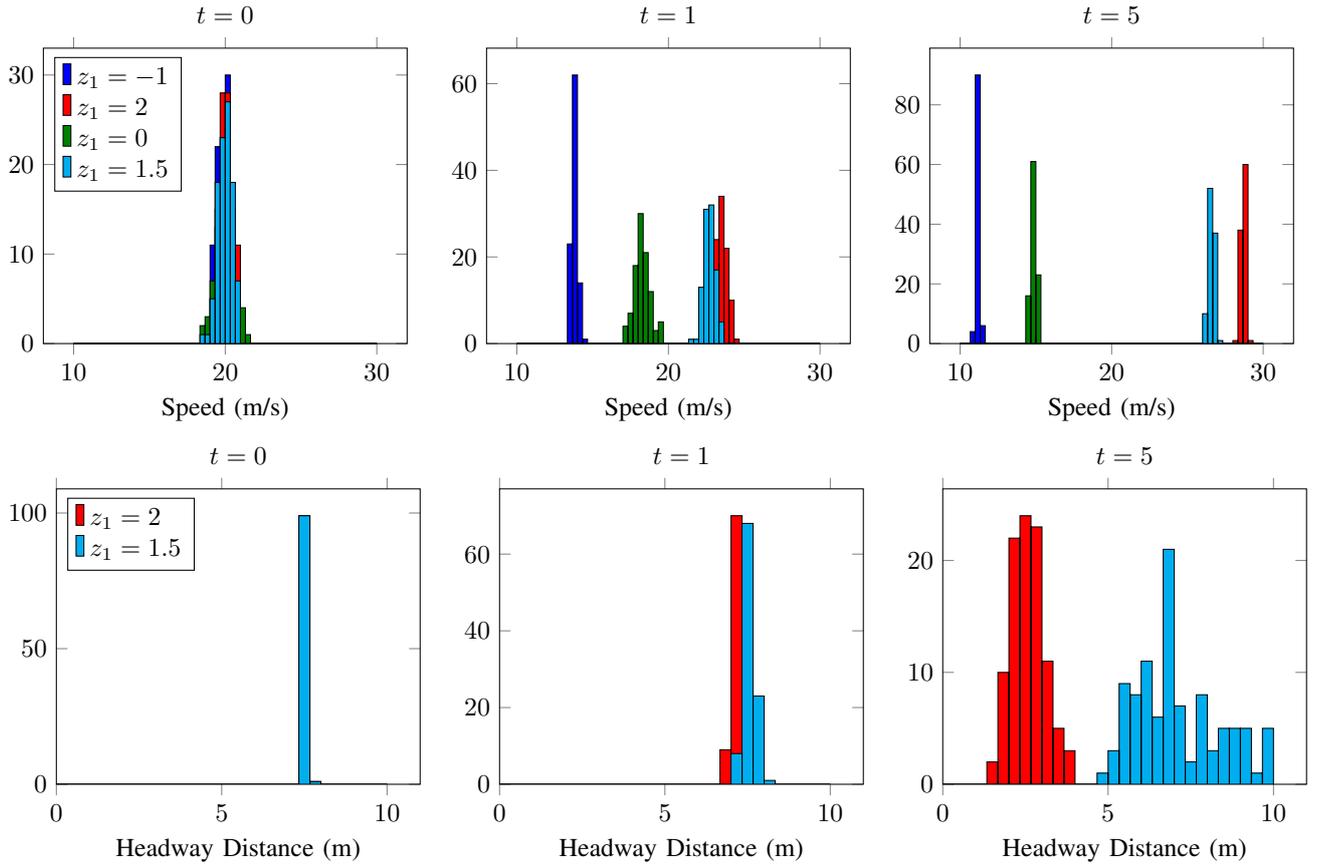

Fig. 6: Distribution over speed and headway values at time horizons of 0, 1, and 5 seconds.

Our policies are optimized with a behavioral cloning objective, and they are susceptible to degrading performance when they find themselves in parts of the state space that are not seen in the training data. While this makes our policies currently unsuitable for simulating real-world driving behavior, future work will combine our approach with reward-based approaches for driver behavior modeling. In particular, our learned encoder can be used to find latent encodings for expert trajectories, which will subsequently allow us to improve our policies through generative adversarial imitation learning. Additionally, while the latent states in this work correspond to driver behavior classes, in future work this same approach could be extended to datasets with latent states corresponding to driver intention or maneuver classes. The code associated with this paper and relevant videos can be found at https://github.com/sisl/latent_driver.

ACKNOWLEDGMENT

This material is based upon work supported by the Ford Motor Company and the National Science Foundation Graduate Research Fellowship Program under Grant No. DGE-114747. The authors would like to thank Alex Kuefler, Tim Wheeler, and Rui Shu for their help and feedback.